\documentclass{article}

\usepackage[utf8]{inputenc} 
\usepackage[T1]{fontenc}    
\usepackage{hyperref}       
\usepackage{url}            
\usepackage{booktabs}       
\usepackage{amsfonts}       
\usepackage{nicefrac}       
\usepackage{microtype}      
\usepackage{graphicx}
\usepackage{doi}
\usepackage{amsmath}
\usepackage{amssymb}
\usepackage[margin=1in]{geometry}
\usepackage[table]{xcolor}
\usepackage{algorithm}
\usepackage{algpseudocode}

\hypersetup{
pdftitle={Decorrelative Network Architecture for Robust Electrocardiogram Classification},
pdfsubject={},
pdfauthor={Christopher Wiedeman, Ge Wang},
pdfkeywords={Deep Learning, Electrocardiogram, Adversarial Attack Stability, Ensemble Learning},
}

\title{Decorrelative Network Architecture for Robust Electrocardiogram Classification}

\author{Christopher Wiedeman$^1$ \and Ge Wang$^2$ \thanks{Correspondence: wang6@rpi.edu}}

\date{
	$^1$Department of Electrical and Computer Systems Engineering\\
	Rensselaer Polytechnic Institute \\ \texttt{wiedec@rpi.edu}\\%
	$^2$Department of Biomedical Engineering\\
	Rensselaer Polytechnic Institute \\ \texttt{wang6@rpi.edu}\\%
}

\begin{document}
\maketitle

\begin{abstract}
Artificial intelligence has made great progress in medical data analysis, but the lack of robustness and trustworthiness has kept these methods from being widely deployed. As it is not possible to train networks that are accurate in all scenarios, models must recognize situations where they cannot operate confidently. Bayesian deep learning methods sample the model parameter space to estimate uncertainty, but these parameters are often subject to the same vulnerabilities, which can be exploited by adversarial attacks. We propose a novel ensemble approach based on feature decorrelation and Fourier partitioning for teaching networks diverse complementary features, reducing the chance of perturbation-based fooling. We test our approach on single and multi-channel electrocardiogram classification, and adapt adversarial training and DVERGE into the Bayesian ensemble framework for comparison. Our results indicate that the combination of decorrelation and Fourier partitioning generally maintains performance on unperturbed data while demonstrating superior robustness and uncertainty estimation on projected gradient descent and smooth adversarial attacks of various magnitudes. Furthermore, our approach does not require expensive optimization with adversarial samples, adding much less compute to the training process than adversarial training or DVERGE. These methods can be applied to other tasks for more robust and trustworthy models.
\end{abstract}

\textbf{Keywords: }Deep Learning, Adversarial Attacks, Bayesian Neural Networks, Electrocardiogram Classification
\break

\section*{Introduction}
The exponential increase of high-dimensional patient datasets and constant demand for personalized healthcare justify the urgent need for artificial intelligence (AI) in medicine. As an excellent example, electrocardiograms (ECG) are commonly used for inpatient monitoring of cardiac conditions, and are now available in smart or implantable devices. While the proper application of continuous ECG monitoring requires further clinical investigation, large scale collection and analysis of ECG, in either inpatient or outpatient populations, has the potential to improve healthcare by monitoring for signs of heart problems or alerting medical services to emergency situations. Deeper analysis of numerous samples is necessary to extract more healthcare-relevant information hidden in these signals. Big data can be leveraged in this instance, but it is infeasible for human clinicians to individually analyze all these recordings, making AI a natural solution to this problem \cite{murat_application_2020, xiao_2022, hong_2020}.

For this purpose, many researchers applied deep learning to ECG classification. The 2017 PhysioNet Challenge is a milestone in this field, where deep neural networks (DNNs) were trained to classify atrial fibrillation from single-lead ECG \cite{clifford_af_2017}. Similarly, the 2018 China Physiological Signal Challenge (CPSC 2018) observed classification of several rhythm abnormalities from 12-lead ECG \cite{china_2018_data}. Top-scoring models can often achieve high classification accuracies on test data, but their interpretability and robustness are major concerns \cite{goodfellow_towards_2018}. Chief among these concerns are adversarial attacks, which have been demonstrated both in machine learning broadly and specific healthcare tasks.

\subsection*{Adversarial Attacks: Background and Characteristics}
Adversarial attacks are small input perturbations that do not change the semantic content yet cause massive errors in a network output; for example, imperceivable noise patterns that, when added to an image, cause a model misclassify the image \cite{szegedy_intriguing_2014}. Given a target model and input, projected gradient descent (PGD) is the most common algorithm for finding adversarial perturbations under an $\ell_\infty$ bound \cite{madry_towards_2019, ren_adversarial_2020}. Various other algorithms exist, including the use of generative adversarial ensembles \cite{song_constructing_2018, xiao_generating_2019, liu_rob-gan_2019}. Impressively, \emph{universal adversarial perturbations} can be crafted to fool a network when added to any sample \cite{moosavi-dezfooli_deepfool_2016, chaubey_universal_2020}.

The understanding of adversarial attacks has rapidly developed over the past several years. Akhtar and Mian wrote a broad survey on adversarial attacks in computer vision \cite{akhtar_threat_2018}. Although adversarial instability may relate to overfitting, DNNs often generalize well to unseen data yet fail on previously seen data that are only slightly altered \cite{goodfellow_explaining_2015}. Furthermore, it has been shown that linear models and other machine learning methods are also vulnerable to adversarial attacks \cite{papernot_transferability_2016}. Early research attributed this phenomenon to lack of data in high-dimensional problems, which leaves large portions of the total ‘data-manifold’ unstable \cite{gilmer_adversarial_2018, dube_high_2018}. Literature also reported a relationship between large local Lipschitz constants (with regards to the loss function) and adversarial instability \cite{qin_adversarial_2019, hein_formal_2017, roth_adversarial_2020}. To our knowledge, the most unifying explanation is the robust features model, where it is shown that data distributions often exhibit statistical patterns that are meaningless to humans but correlate well with different classes \cite{ilyas_adversarial_2019}. From a human’s perspective, these patterns are arbitrary and easily perturbed, but since models are trained to only maximize distributional accuracy, they have no reason to prioritize human-favored features over these patterns.

Training models for defending against adversarial attacks remains an open problem, affecting nearly every application of machine learning. Early attempts at defense methods by obfuscating the loss gradient were found to beat only weak attackers, proving ineffective for sophisticated attackers \cite{athalye_obfuscated_2018, carlini_adversarial_2017, uesato_adversarial_2018, carlini_defensive_2016, papernot_distillation_2016}. Adversarial training, in which a model is iteratively trained on strong adversarial samples, has shown the best results in terms of adversarial robustness \cite{madry_towards_2019, kurakin_adversarial_2017-1}. However, the network size and computational time required is considerable for small problems, and improving adversarial robustness appears to sacrifice performance on clean data \cite{yin_fourier_2019}. Satisfactory performance on larger problems has not been achieved due to these limitations.

Another troubling, well-documented characteristic of these attacks is their \emph{transferability}: models trained on the same task will often be fooled by the same attacks, despite having different parameters \cite{szegedy_intriguing_2014, papernot_transferability_2016, tramer_space_2017}. This phenomenon is largely congruent with the robust features model, since these models are likely learning the same useful, but non-robust features. Nevertheless, transferability makes black-box attacks viable, where a malicious attacker does not need detailed knowledge of the target model.

Han, et al. has shown that models trained for ECG classification are concerningly susceptible to natural-looking adversarial attacks \cite{han_deep_2020}. In short, the authors observed that traditional $\ell_\infty$-bounded PGD attacks produce square-wave artifacts that are not physiologically plausible in ECG; to rectify this, the perturbation space was modified by applying Gaussian smoothing kernels in the attack objective, rendering plausible yet still highly effective adversarial samples.

\subsection*{Electrocardiography Background}
ECG is a front-line, noninvasive tool for monitoring heart health. Skin electrodes measure electrical signals originating from the heart; the behavior of these signals over time correspond to various events during the cardiac cycle. A healthy rhythm consists of a \emph{P wave, QRS complex}, and \emph{T wave}, which correspond to atrial depolarization, ventricular depolarization, and ventricular repolarization, respectively. Clinical ECGs have traditionally used 12 or 5-7 (Holter) leads, but single-channel ECGs have become more prevalent for continuous monitoring \cite{merdjanovska_comprehensive_ecg_2022}.  These devices are either external or implantable loop recorders (ILRs) designed to record for multiple years.

A variety of downstream analytical tasks are associated with ECG, including biometric identification, respiratory estimation, emotional monitoring, and even fetal heartbeat monitoring. However, the most common application is detection of various arrythmias, which can indicate disorders or disease risk \cite{merdjanovska_comprehensive_ecg_2022}. Atrial fibrillation (AFib), or an abnormally rapid atrial firing rate, is commonly assessed (such as in the 2017 PhysioNet Challenge \cite{clifford_af_2017}), but many arrythmic classses exist, including left or right bundle branch block, premature atrial or ventricular contraction, ventricular fibrillation,  tachycardia, and myocardial infarction (heart attack) itself. Certain classes require immediate and serious medical intervention while others, such as AFib, are not immediately harmful but could still indicate risk of disease. The clinical utility of AFib detection is still an active area of investigation: a systematic review found AFib to be associated with increased risk of myocardial infarction in patients without coronary heart disease and increased risk of all-cause mortality and heart failure in all patients, implying value in detecting it \cite{ruddox_afib_infarc_2017}. On the other hand, a randomized control trial using ILR in patients with at least one risk factor for stroke concluded that continuously monitoring for AFib in this population did not reduce risk of stroke \cite{svendsen_loop_2021}. This suggests that detecting AFib early may not provide additional information for managing stroke in patients that are already known to be at risk of the event. Nevertheless, ECG and its subsequent analysis is widely applied and investigated for its implications on patient cardiovascular health.

Clinician review is often required for ECG analysis; this process is resource-consuming, especially in the case of continuous monitoring or large patient samples. Thus, automatic classification of these signals is desirable, but simple rule-based classifications often fail to generalize due to data heterogeneity between patients and the non-stationary nature of the signal within patients. As such, researchers have turned to data driven techniques and machine learning to build ECG classification models. Convolutional neural networks (CNNs) have been the most dominant architecture for ECG arrythmia classification, but deep belief networks, recurrent neural networks, long short-term memory, and gated recurrent units have all been investigated for the same task \cite{ebrahimi_review_2020}. For extensive background, Merdjanovska, et al. provides a comprehensive review of applications, public datasets, and deep learning research for ECG, and Ebrahimi, et al. further surveys common deep learning architectures for ECG \cite{merdjanovska_comprehensive_ecg_2022, ebrahimi_review_2020}.

\subsection*{Uncertainty Estimation in Healthcare Applications}
\begin{figure*}
\centering
\includegraphics[width=16.0cm]{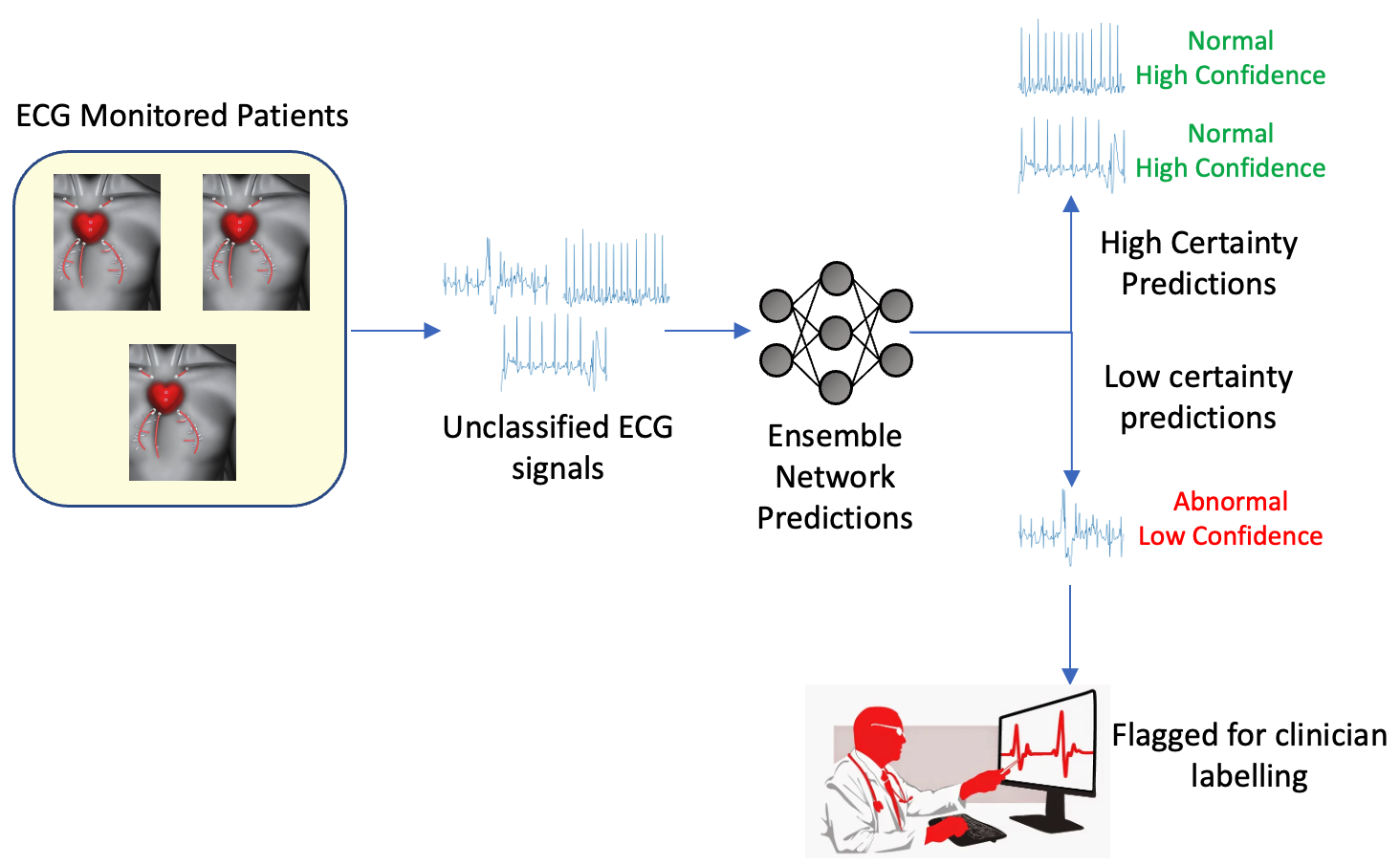}
\caption{Example of proposed AI augmented clinical workflow for monitoring ECG signals in a patient population. Data are first processed by a deep learning model, which infers a class for each signal (e.g., healthy or diseased) and judges the confidence of each inference. Signals classified with low confidence are reviewed by human experts.}\label{fig:workflow}
\end{figure*}

As misdiagnosis in healthcare contexts can cause serious harm, the standard of trust required for AI to operate in this space is high. Rather than replacing clinicians, we envision AI tools augmenting clinical workflows by monitoring inputs over a large population, flagging alarming or low-confidence instances for human observation. Figure \ref{fig:workflow} broadly illustrates this scenario, where AI could allow a few experts to analyze ECG signals from a large patient population, continuously or transiently monitored. To achieve this synergy, models must be capable of gauging their own confidence, recognizing conditions where they can and cannot perform well \cite{elul_2021}. Bayesian deep learning (BDL) is a promising field that models the parameters of a DNN as a distribution rather than a point estimation; sampling this distribution at inference time then allows one to estimate model certainty in an inference \cite{wilson_case_2020, wilson_bayesian_2020}. Approaches for approximating and sampling the parameter distribution, including variational inference and Markov Chain Monte Carlo with Hamiltonian Dynamics, are often difficult to scale to large spaces \cite{Graves2011PracticalVI, Neal_1992}. One simple approach is to train an ensemble of networks for the same task, with each network acting as a sample of the parameter space \cite{balaji_2017}. However, this approach does not guarantee robustness: adversarial attacks in particular are known to transfer between different models because these models (even with vastly different parameters) often learn the same unstable features. Furthermore, in high dimensional problems with large parameter spaces, training, storing, and running inferences from numerous models quickly becomes infeasible. As such, the goal in training such ensembles should be to achieve adequate robustness and feature diversity with a small number of models.

\subsection*{Motivation: Diversifying Features in Deep Ensembles}
To our knowledge, prior work on adversarial robustness has primarily quantified either white-box accuracy or black-box transferability, but have not evaluated uncertainty via BDL. Furthermore, works in this field primarily test methods on lower dimension datasets, such as MNIST or CIFAR10. Our goal is to efficiently train small but diverse deep ensembles capable of gauging uncertainty in worst-case scenarios, i.e., adversarial attacks. We contextualize this in the aforementioned ECG classification, a problem that is much higher in dimension. Adversarial training as a means of diversifying an ensemble is explored, and we also introduce two novel diversification methods that do not require adversarial sample computations, adding almost no overhead to the regular training process. 

According to the robust features model, simply training networks with different parameters in isolation does not achieve adversarial robustness, as networks trained under the same conditions tend to converge toward the same learned features and vulnerabilities \cite{ilyas_adversarial_2019, wiedeman_disrupting_2022}. As such, rather than achieving diversity in the parameter space, we turn the conversation to diversity in the feature space. A mechanism for incentivizing networks to learn different features is necessary. To this end, Yang, et al. conceived DVERGE, which diversifies the learned features and adversarial weaknesses in a classification ensemble \cite{yang_dverge_2020}. However, this method requires full or partial computations of adversarial samples and round-robin style training of networks, which adds considerable compute. In this work, we experiment with ensemble diversification methods that are based on adversarial gradients and other methods that are agnostic to these calculations.

\subsubsection*{Adapting Adversarial Training for Ensembles}
Adversarial training is the best known defense against adversarial attacks, and essentially consists of training a network on adversarial samples \cite{madry_towards_2019}. Unfortunately, adversarial training is computationally expensive and reduces accuracy on natural data \cite{yin_fourier_2019}. 

Conventional adversarial training is the best known defense against adversarial attacks, but it does not detect attacks or quantify uncertainty; rather, it attempts to make a single network more robust to attacks. Furthermore, it only achieves satisfactory performance on small problems using large networks to fit more complex decision boundaries \cite{madry_towards_2019}. As such, we adapt \emph{adversarial trained ensembles}, in which individual networks are adversarially trained for additional time after natural training. Consequently, we are able to study adversarial training in the Bayesian ensemble framework, and observe how vulnerabilities may be diversified between models.

\subsubsection*{Alternative Feature Diversification Methods}
We propose two distinct methods for diversifying learned features, and test these against adversarial ECG attacks in \cite{han_deep_2020}. The first method, linear feature decorrelation, is based on previous work \cite{wiedeman_disrupting_2022}, which not only found a strong linear correlation between in the latent space of networks trained on the same task, but also found that adding a loss term to reduce the linear correlation greatly decreases the transferability of adversarial attacks. However, the decorrelation process proposed in that work is expensive, as it requires large batch sizes and parallel training of networks. We modify this decorrelation process to make it scalable to larger problems. The second method which we refer to as Fourier partitioning, is heuristically simpler, employing linear time-invariant filters to partition the input space by frequency, forcing networks to learn features in different frequency bands. This method is inspired by recently discovered connections between the Fourier space and adversarial vulnerability, which not only demonstrated that neural networks can make accurate inferences by relying only on low or high-frequency characteristics but also that most robustifying training methods only shift a network’s sensitivity to different frequency bands \cite{yin_fourier_2019}. As such, we find that a crude but efficient way to teach networks different features is to partition the original inputs by frequency, feeding data in different bands to different networks and integrate their outputs via ensemble learning.

\section*{Results}
\subsection*{Overview}
Two ECG datasets are used for all experiments: the 2017 PhysioNet challenge data (single-channel, four classes) \cite{clifford_af_2017} and the 2018 China Physiological Signal Challenge (CPSC) data (twelve-channel, nine classes) \cite{china_2018_data}.  The following ensemble training strategies were tested:
\begin{itemize}
    \item \textbf{baseline}: Conventionally trained ensemble, where each model is identically and independently trained.
    \item \textbf{dec}: Ensemble trained with the proposed linear feature decorrelation to diversify the model features.
    \item \textbf{part}: Ensemble trained using the proposed Fourier partitioning scheme.
    \item \textbf{adv}: A baseline ensemble that undergoes additional ensemble adversarial training.
    \item \textbf{dec+part}: Ensemble that employs both linear feature decorrelation and Fourier partitioning.
    \item \textbf{dec+adv}: A decorrelated ensemble that undergoes additional ensemble adversarial training.
    \item \textbf{dverge}: A baseline ensemble that undergoes additional DVERGE training.
\end{itemize}

Ensembles produce multiple inferences, which can be processed in various ways to gauge epistemic and aleatoric uncertainty\cite{gal2016dropout, kendall_2017}. Here, we adopt a normalized uncertainty approach from \cite{mobiny_2021}, which calculates a normalized measure of certainty $I_{norm}$ based on the mutual information between the sample and the model parameters (see Methods: Ensemble Training and Inference). 

\begin{figure}
\centering
\includegraphics[width=16.0cm]{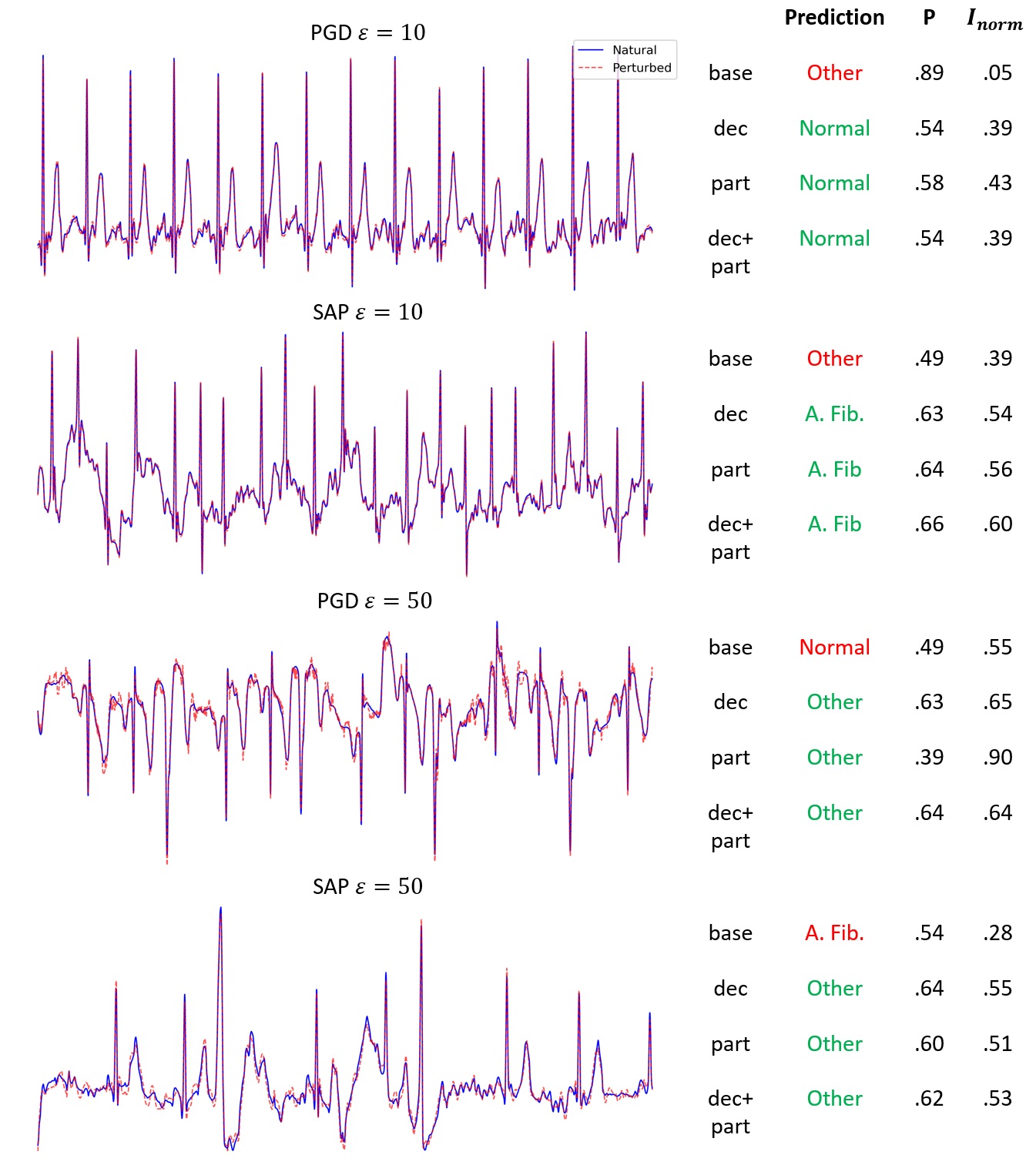}
\caption{Sample results from the Physionet 2017 data. Left: Examples of projected gradient descent (PGD) and smooth adversarial perturbations (SAP) in the ECG dataset. Right: correct (green) or incorrect (red) aggregate inferences of each ensemble network (normal rhythm, atrial fibrillation, other rhythm, or noise) along with the inferred class probability P and normalized uncertainty score $I_{norm}$.}
\label{fig:ecg_samples}
\end{figure}

We test each ensemble using validation data perturbed by both PGD and physiologically feasible SAP attacks of varying magnitude $\varepsilon$, targeting the first model in each ensemble \cite{han_deep_2020}. Figure \ref{fig:ecg_samples} displays several example attacks from the PhysioNet 2017 data along with inferences, probability, and uncertainty values outputted by the baseline, dec, part, and dec+part ensembles. 

Notably, we found that implementing the decorrelation step only slightly increased training time: conventional training took about 352 and 358 minutes per model on average for PhysioNet and CPSC, respectively; decorrelation only added about 10 minutes on average to this time in both cases (see Methods: Experimental Details for training parameters). Fourier partitioning did not noticeably increase training time. 

\subsection*{Uncertainty and Accuracy Performance}
To classify an ensemble inference as either certain or uncertain, a threshold $I_T \in [0,1]$, can be applied to differentiate certain $I_{norm}\leq I_T$ and uncertain $I_{norm}>I_T$ predictions. A robust model is generally correct when it is certain and uncertain when it is incorrect. Thus, in addition to inference accuracy, we also adopt the following three evaluation metrics from \cite{mobiny_2021}:

\begin{itemize}
  \item Correct-certain ratio $R_{cc}(I_T)=P_{I_T}(correct|certain)$: Probability the model inference is correct when it is certain.
  \item Incorrect-uncertain ratio $R_{iu}(I_T)=P_{I_T}(uncertain|incorrect)$: Probability the model is uncertain when it is incorrect.
  \item Uncertainty Accuracy $UA(I_T)=P_{I_T}(correct \cap certain \bigcup incorrect \cap  uncertain)$: Probability of a desired outcome (either correct and certain of uncertain and incorrect).
\end{itemize}

All three measures depend on the variable uncertainty threshold $I_T$. Thus, similar to a binary classifier, the overall efficacy can be found by integrating the measure as a function of $I_T \in [0,1]$ (i.e., finding the area under the curve, where larger values are more desirable).

Table \ref{table:results} summarizes the average prediction accuracy and areas under the curve (AUCs) for the correct-certain ratio, incorrect-uncertain ratio, and uncertainty accuracy for natural, PGD, and SAP adversarial datasets for the PhysioNet 2017 experiments. Similarly, Table \ref{table:results_china} reports the same metrics for the CPSC 2018 experiments. Some initial observations from these numbers are as follows: 1) For the PhysioNet 2017 experiments, dec, part, and dec+part generally have comparable or superior metrics to the baseline on natural ($\varepsilon=0$) samples, but achieve better performance on stronger $\varepsilon=50, 75, 100$ PGD and SAP attacks. Adversarial training (adv) improves performance on all adversarial attacks, but the combination of dec+adv generally leads to better performance in these instances, while dverge does not seem to improve from the baseline in this instance. 2) On the CPSC 2018 experiments, adv and dec+adv achieve improved robustness on perturbed data but sacrifice considerable accuracy on natural samples. dverge experiences a mild reduction in natural performance for a moderate performance increase on perturbed data, although this advantage diminishes with higher magnitude attacks. The combination of dec+part poses better performance on perturbed data relative to the baseline without sacrificing performance on natural samples. 

\begin{table}
\centering
\begin{tabular}{ c | c | c c c c | c c c c  }
\hline
 &  & \multicolumn{4}{c}{Attack Strength $\varepsilon$ (PGD)} & \multicolumn{4}{c}{Attack Strength $\varepsilon$ (SAP)} \\ 
 & 0 & 10 & 50 & 75 & 100 & 10 & 50 & 75 & 100 \\ 
\hline
Accuracy (\%) \\ 
\hline
baseline & 83.82 & 72.45 & 10.90 & 5.63 & 4.10 & 74.91 & 16.18 & 8.32 & 5.39 \\ 
dec & 84.41 & 72.57 & 26.61 & 13.95 & 9.50 & 74.21 & 27.78 & 14.77 & 7.50 \\ 
part & \textbf{86.75} & 70.46 & 32.12 & 23.56 & 19.81 & 72.33 & 38.45 & 25.67 & 19.58 \\ 
dec+part & 85.58 & 73.86 & 54.51 & \textbf{44.55} & \textbf{36.34} & 75.38 & 57.80 & 46.66 & \textbf{36.58} \\ 
adv & 83.35 & \textbf{79.72} & 58.15 & 32.59 & 15.01 & \textbf{80.42} & \textbf{66.71} & 45.25 & 21.45 \\ 
dverge & 83.59 & 69.52 & 10.67 & 5.39 & 2.81 & 71.75 & 14.30 & 7.27 & 4.34 \\ 
dec+adv & 80.07 & 77.02 & \textbf{60.14} & 43.96 & 26.73 & 77.96 & 64.83 & \textbf{50.29} & 33.53 \\ 
\hline
$R_{cc}$ (\% AUC) \\ 
\hline
baseline & 85.46 & 74.17 & 4.77 & 2.00 & 1.38 & 76.63 & 8.23 & 3.17 & 1.88 \\ 
dec & 86.18 & 76.74 & 16.22 & 7.29 & 4.95 & 78.35 & 17.63 & 7.08 & 3.34 \\ 
part & \textbf{88.18} & 73.10 & 18.64 & 13.40 & 11.23 & 75.80 & 23.20 & 15.05 & 11.76 \\ 
dec+part & 87.64 & 77.50 & 27.76 & 22.72 & 18.76 & 79.29 & 30.02 & 23.89 & 19.09 \\ 
adv & 85.08 & \textbf{81.72} & 61.40 & 34.90 & 13.74 & \textbf{82.42} & \textbf{69.24} & 47.09 & 21.57 \\ 
dverge & 85.00 & 69.55 & 4.03 & 2.18 & 1.25 & 71.84 & 5.45 & 2.69 & 1.84 \\ 
dec+adv & 81.20 & 78.33 & \textbf{63.27} & \textbf{47.63} & \textbf{26.72} & 79.19 & 67.55 & \textbf{54.31} & \textbf{35.65} \\ 
\hline
$R_{iu}$ (\% AUC) \\ 
\hline
baseline & 13.77 & 19.57 & 34.20 & 37.69 & 35.76 & 18.52 & 37.45 & 44.23 & 44.91 \\ 
dec & 15.91 & 29.62 & 44.57 & 41.51 & 38.71 & 28.87 & 46.06 & 42.41 & 39.02 \\ 
part & 15.27 & 28.56 & \textbf{46.64} & 46.18 & 43.85 & 28.53 & \textbf{50.02} & \textbf{51.57} & 51.87 \\ 
dec+part & \textbf{19.57} & \textbf{29.69} & 42.53 & \textbf{49.03} & \textbf{51.06} & \textbf{30.19} & 43.36 & 50.87 & \textbf{53.58} \\ 
adv & 16.97 & 18.65 & 24.37 & 22.07 & 16.17 & 18.57 & 24.49 & 27.19 & 20.53 \\ 
dverge & 12.68 & 13.85 & 15.25 & 14.73 & 15.99 & 12.88 & 15.97 & 19.59 & 22.44 \\ 
dec+adv & 10.58 & 11.60 & 16.81 & 21.80 & 23.39 & 11.47 & 16.12 & 21.35 & 23.74 \\ 
\hline
$UA$ (\% AUC) \\ 
\hline
baseline & 81.65 & 67.27 & 33.73 & 36.94 & 35.32 & 69.92 & 36.56 & 42.51 & 43.68 \\ 
dec & 81.31 & 68.22 & 44.01 & 41.44 & 39.01 & 70.33 & 45.08 & 41.70 & 38.87 \\ 
part & \textbf{82.74} & 62.36 & 45.81 & 45.62 & 44.10 & 65.75 & 47.35 & 49.19 & \textbf{50.23} \\ 
dec+part & 81.07 & 66.66 & 40.92 & 44.42 & \textbf{46.80} & 68.69 & 40.78 & 45.06 & 48.20 \\ 
adv & 78.04 & 74.60 & 57.12 & 40.86 & 24.92 & 75.42 & 61.34 & 47.03 & 31.54 \\ 
dverge & 80.35 & 64.00 & 17.32 & 15.99 & 16.70 & 65.99 & 18.60 & 20.61 & 23.11 \\ 
dec+adv & 77.43 & \textbf{75.19} & \textbf{61.68} & \textbf{49.73} & 37.68 & \textbf{75.93} & \textbf{65.22} & \textbf{54.52} & 42.60 \\ 
\hline
\end{tabular} 
\caption{Performance metrics on PGD and SAP adversarial samples of varying magnitudes on the PhysioNet 2017 dataset. $R_{cc}$: correct-certain ratio, $R_{iu}$: incorrect-uncertain ratio, and $UA$: uncertainty accuracy.}
\label{table:results}
\end{table}

\begin{table}
\centering
\begin{tabular}{ c | c | c c c c | c c c c  }
\hline
 &  & \multicolumn{4}{c}{Attack Strength $\varepsilon$ (PGD)} & \multicolumn{4}{c}{Attack Strength $\varepsilon$ (SAP)} \\ 
 & 0 & .025 & .05 & .15 & .175 & .025 & .05 & .15 & .175 \\ 
\hline
Accuracy (\%) \\ 
\hline
baseline & \textbf{80.28} & 41.47 & 23.94 & 3.60 & 3.13 & 41.00 & 23.16 & 2.82 & 1.88 \\ 
dec & 79.66 & 38.50 & 21.91 & 2.03 & 1.41 & 40.22 & 20.50 & 0.78 & 0.16 \\ 
part & 79.81 & 35.68 & 16.28 & 3.44 & 2.82 & 37.25 & 9.86 & 0.94 & 1.41 \\ 
dec+part & \textbf{80.28} & 44.76 & 31.77 & 12.68 & 8.45 & 46.79 & 33.33 & \textbf{18.78} & \textbf{17.53} \\ 
dverge & 78.40 & 47.26 & 30.99 & 12.36 & 9.08 & 48.83 & 31.92 & 10.80 & 5.63 \\ 
adv & 58.37 & \textbf{49.61} & 37.25 & \textbf{16.90} & \textbf{13.93} & \textbf{50.86} & \textbf{38.03} & 15.49 & 11.58 \\ 
dec+adv & 58.06 & 46.48 & \textbf{37.40} & 11.89 & 9.08 & 47.10 & \textbf{38.03} & 14.40 & 9.08 \\ 
\hline
$R_{cc}$ (\% AUC) \\ 
\hline
baseline & 84.92 & 33.09 & 7.16 & 0.49 & 0.36 & 32.13 & 6.16 & 0.18 & 0.15 \\ 
dec & 83.90 & 33.64 & 6.85 & 0.55 & 0.32 & 31.96 & 5.96 & 0.17 & 0.05 \\ 
part & 84.13 & 48.71 & 9.68 & 1.30 & 0.85 & 47.46 & 4.51 & 0.58 & 0.51 \\ 
dec+part & \textbf{85.54} & 53.50 & 13.81 & 4.84 & 2.66 & 51.85 & 12.32 & 6.89 & 5.28 \\ 
dverge & 82.60 & 39.64 & 9.82 & 2.02 & 1.03 & 40.37 & 8.75 & 1.55 & 0.38 \\ 
adv & 70.98 & \textbf{56.13} & 31.22 & \textbf{9.21} & \textbf{7.19} & \textbf{57.57} & 33.19 & \textbf{7.50} & \textbf{5.81} \\ 
dec+adv & 65.68 & 54.28 & \textbf{42.92} & 8.89 & 5.28 & 54.79 & \textbf{43.81} & 7.11 & 4.70 \\ 
\hline
$R_{iu}$ (\% AUC) \\ 
\hline
baseline & 30.01 & 32.03 & 32.89 & 27.75 & 26.73 & 34.41 & 38.46 & 42.40 & 45.67 \\ 
dec & 28.02 & 34.52 & 34.35 & 30.18 & 32.05 & 36.94 & 37.30 & 35.10 & 42.43 \\ 
part & 31.47 & 48.33 & 57.04 & 52.14 & 50.30 & 48.08 & 60.47 & \textbf{70.94} & \textbf{72.94} \\ 
dec+part & 35.65 & 50.75 & \textbf{61.70} & \textbf{61.71} & \textbf{61.59} & 52.28 & \textbf{63.92} & 68.91 & 71.33 \\ 
dverge & 26.13 & 30.93 & 32.98 & 35.74 & 35.46 & 31.40 & 37.26 & 39.37 & 38.64 \\ 
adv & \textbf{59.11} & \textbf{58.11} & 58.15 & 60.71 & 61.17 & \textbf{58.00} & 58.34 & 59.84 & 63.95 \\ 
dec+adv & 29.77 & 28.97 & 27.53 & 30.45 & 32.70 & 28.86 & 27.91 & 41.75 & 41.07 \\ 
\hline
$UA$ (\% AUC) \\ 
\hline
baseline & \textbf{77.91} & 41.90 & 30.44 & 27.20 & 26.23 & 42.24 & 33.89 & 41.37 & 44.94 \\ 
dec & 75.82 & 43.38 & 32.81 & 30.08 & 31.90 & 42.49 & 34.99 & 34.97 & 42.41 \\ 
part & 70.20 & 53.04 & \textbf{54.28} & 51.48 & 49.64 & 52.07 & \textbf{58.03} & \textbf{70.58} & \textbf{72.25} \\ 
dec+part & 71.54 & 52.30 & 51.32 & \textbf{57.48} & \textbf{58.52} & 51.21 & 51.82 & 60.38 & 62.21 \\ 
dverge & 76.31 & 42.88 & 30.22 & 33.02 & 33.12 & 42.92 & 31.80 & 36.43 & 36.79 \\ 
adv & 51.38 & 50.74 & 51.80 & 56.61 & 57.50 & 50.58 & 51.88 & 55.88 & 60.51 \\ 
dec+adv & 60.84 & \textbf{54.51} & 47.87 & 34.03 & 34.26 & \textbf{54.72} & 48.58 & 41.82 & 41.28 \\ 
\hline
\end{tabular} 
\caption{Performance metrics on PGD and SAP adversarial samples of varying magnitudes on the CPSC 2018 dataset. $R_{cc}$: correct-certain ratio, $R_{iu}$: incorrect-uncertain ratio, and $UA$: uncertainty accuracy.}
\label{table:results_china}
\end{table}

\subsection*{Uncertainty Difference Between Incorrect and Correct Predictions}
Since uncertainty $I_{norm}$ should be a useful discriminative feature for distinguishing correct and incorrect predictions, it is desirable that incorrect predictions, on average, output higher uncertainty than correct predictions. Thus, we simply define the average difference in uncertainty between incorrectly and correctly classified samples:

\begin{align*}
\Delta I_{norm} = \mathbb{E}[I_{norm}|incorrect] - \mathbb{E}[I_{norm}|correct]
\end{align*}

\begin{figure}
\centering
\includegraphics[width=14.0cm]{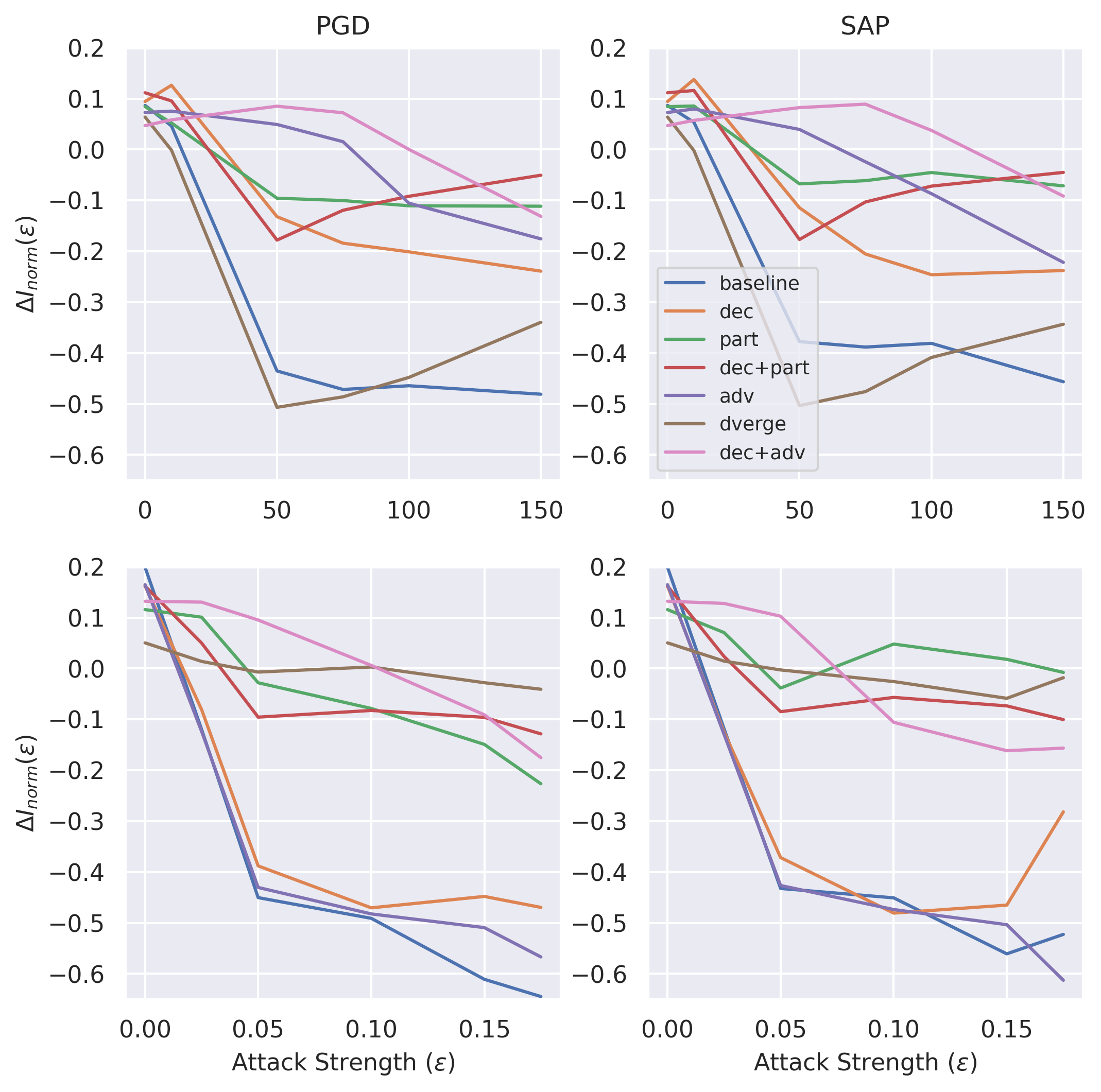}
\caption{Difference in average normalized uncertainty between incorrect and correct samples (higher is better) on PGD (left) and SAP (right) adversarial samples with respect to attack magnitude $\varepsilon$. Top: PhysioNet 2017. Bottom: CPSC 2018}
\label{fig:uncertainty_plots}
\end{figure}

Figure \ref{fig:uncertainty_plots} compares this  $\Delta I_{norm}$ for all experiments and ensembles as a function of attack strength $\varepsilon$. In all instances, $\Delta I_{norm}$ decreases sharply for the baseline ensemble as perturbation strength increases. dverge follows this same trend on the PhysioNet data but maintains better robustness on the CPSC data. part, dec, dec+part, and dec+adv generally maintain higher $\Delta I_{norm}$ on perturbed data in both cases.

\subsection*{Performance on Partially Attacked Dataset}
In a clinical setting, deep models should be used to augment clinical workflows, as shown in Figure \ref{fig:workflow}. When the amount of data needing analysis outstrips the available time of qualified clinicians, deep ensembles can initially assess all inputs and defer the most uncertain samples to human readers. This begs the question: \textit{If a deep ensemble is budgeted a certain number of cases that it can refer to human experts, then how many cases would still be misclassified}? To investigate this, we run the following hypothetical experiment: 1) A dataset of all natural samples and a partially perturbed dataset are drawn. In the partially perturbed dataset, only 50\% of the data are unperturbed, and 25\%, 15\%, and 10\%, of the data have $\varepsilon = 10, 50, 75$ and $\varepsilon = 0.025, 0.05, 0.15$ perturbations for the PhysioNet and CPSC data, respectively. 2) Each ensemble evaluates all samples, ordering the inferences from most to least confident. 3) Starting with the most confident samples, varying amounts of samples would defer to the deep model's classification, while all other (less confident) samples would be correctly classified, presumably reviewed by clinicians. Note that this experiment is not meant to exactly reflect the actions and metrics of a true clinical workflow; rather, it is to investigate and compare the potential benefit of the aforementioned deep ensemble methods in augmenting human workers, particularly in the face of adversarially corrupted data.

\begin{figure}
\centering
\includegraphics[width=14.0cm]{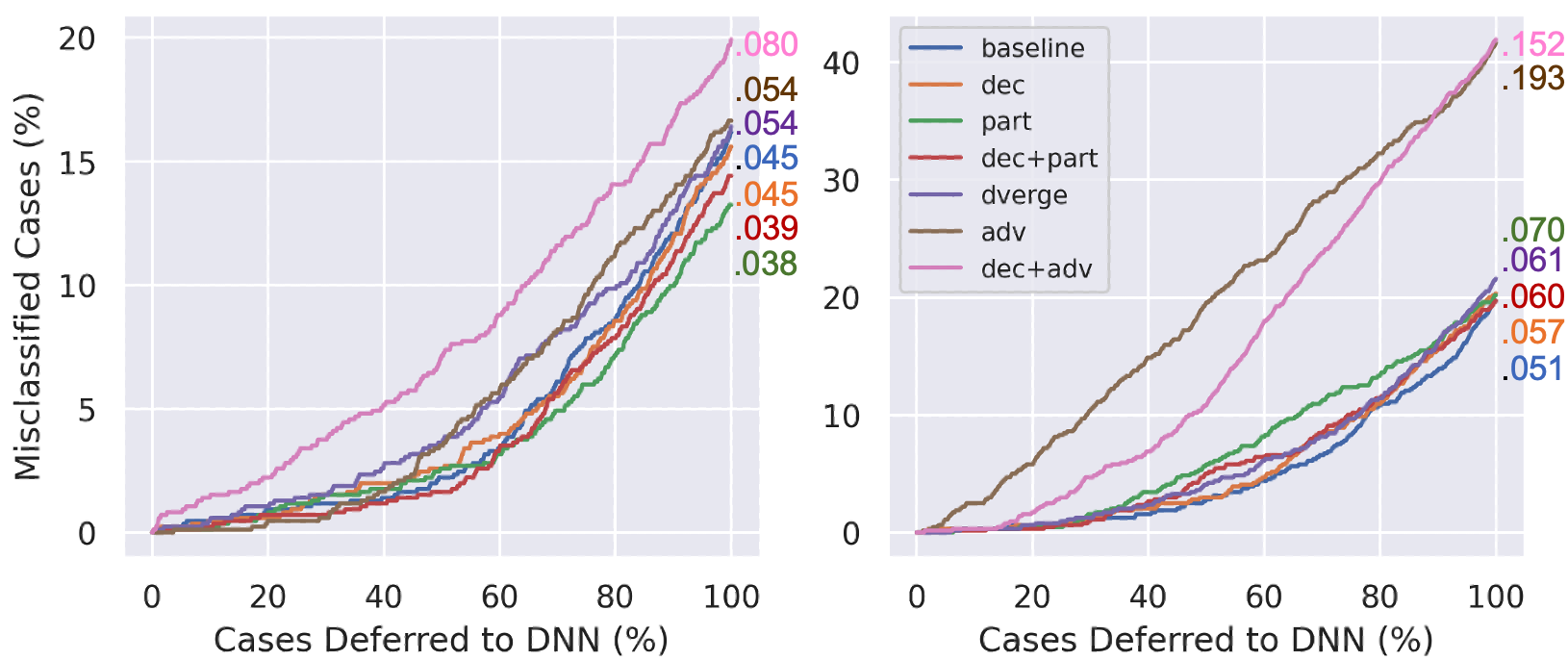}
\caption{\% of misclassified cases with respect to \% of cases deferred to the deep ensemble in the natural only dataset experiments for PhysioNet 2017 (left) and CPSC 2018 (right). Numbers are AUCs (lower is better).}
\label{fig:plot_curve_natural}
\end{figure}

\begin{figure}
\centering
\includegraphics[width=14.0cm]{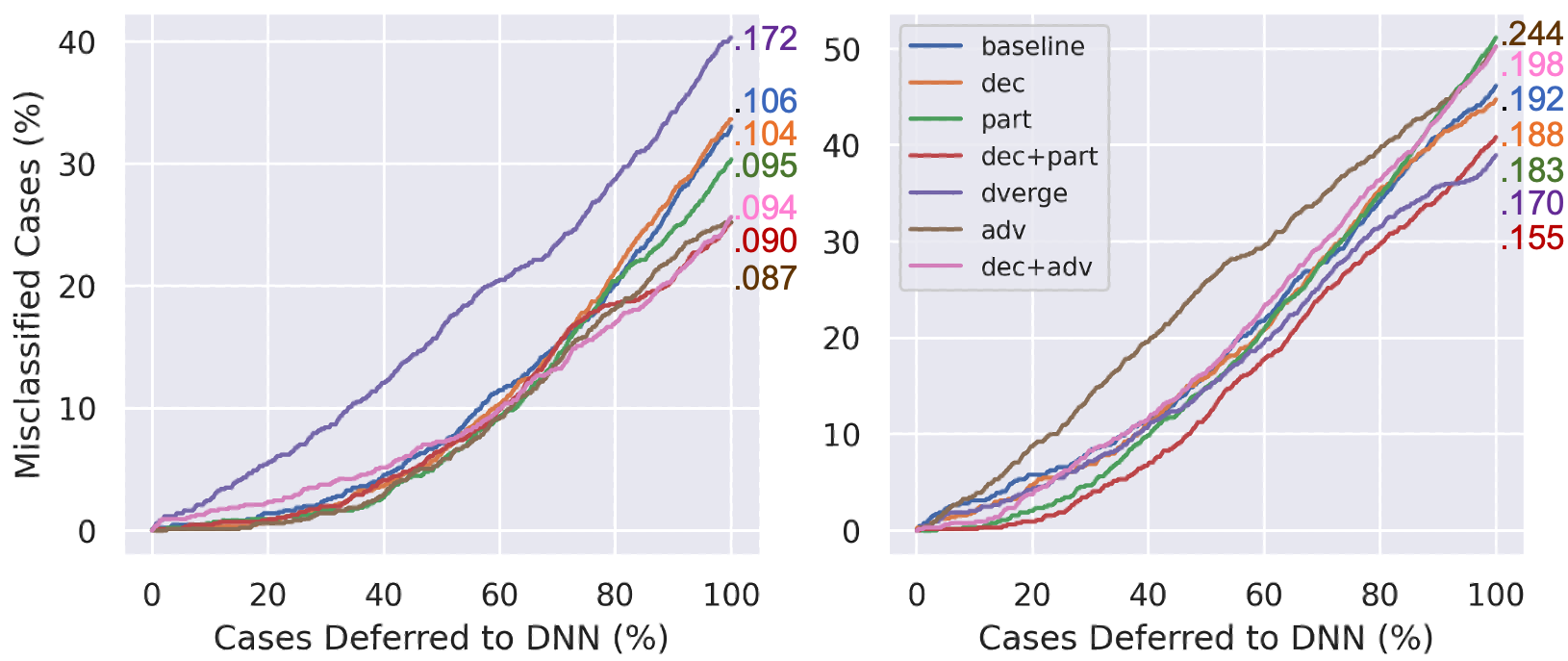}
\caption{\% of misclassified cases with respect to \% of cases deferred to the deep ensemble in the partially attacked dataset experiments for PhysioNet 2017 (left) and CPSC 2018 (right). Numbers are AUCs (lower is better).}
\label{fig:plot_curve}
\end{figure}

Figures \ref{fig:plot_curve_natural} and \ref{fig:plot_curve} plot the percentage of misclassified instances in the sample as a function of the percentage of cases referred to the deep ensemble for the natural and partially perturbed datasets, respectively. Additionally, the total area-under-curves are shown, with lower values being better in this instance. Initially, one can see that adv and adv+dec perform poorly on the natural datasets, particularly the CPSC 2018 data. dec, part, and dec+part perform better than the baseline on the partially perturbed datasets while still performing comparably or better than the baseline on the natural PhysioNet 2017 dataset.

\section*{Discussions}
Table \ref{table:results} reflects the accuracy and uncertainty scores of tested ensembles on the PhysioNet 2017 data. Overall, it can be seen that decorrelation and partioning does not negatively impact natural accuracy in this instance: in fact, overall ensemble accuracy, $R_{cc}$ and $R_{iu}$  are marginally higher on natural data in the dec, part, and dec+part compared to baseline. Furthermore, all ensembles, including the baseline, are more robust to small magnitude perturbations: \cite{han_deep_2020} reported that their $\varepsilon = 10$ SAP attacks fooled a single network 74\% of the time, but our results show only a $26.09\%$ error rate for the baseline ensemble under these conditions. This indicates that in this instance, the network size is large enough relative to the input data dimension to reduce the adversarial transferability in lower magnitude attacks, even without diversifying measures \cite{madry_towards_2019}. However, this initial robustness plummets in the face of more challenging, larger magnitude attacks, as seen in Table \ref{table:results}. The use of decorrelation and partitioning both seem to have positive effects on the ensemble robustness against higher magnitude attacks, with dec+part outperforming even the more expensive adversarially trained ensemble on accuracy and $R_{iu}$ on $\varepsilon=75,100$ PGD and SAP attacks. $UA$ in particular is highest with the combination of decorrelation and adversarial training on all but highest magnitude adversarial attacks. These observations all suggest improved network diversification with both feature decorrelation and Fourier partitioning.

Table \ref{table:results} also indicated little to no benefit from DVERGE training on the PhysioNet 2017 data, as dverge has slightly reduced natural accuracy compared to baseline while showing no general improvement on any metrics for any attack strength. We theorize that this lack of improvement may be due to the low number (4) of discrete classes in the problem. DVERGE creates new samples by distilling the non-robust features of one randomly drawn sample onto another \cite{yang_dverge_2020}. However, if these two samples belong to the same class (which is more likely to occur when there are few classes or class imbalances), then the features distilled from one sample may already be similar to the other, negating any benefit. To support this explanation, DVERGE shows some robustness benefit on the CPSC 2018 data (Table \ref{table:results_china}), which has 9 classes. Consequently, the number of classes and class balance should be considered when implementing DVERGE.

One can see that while adversarial training minimally affects natural performance on the PhysioNet data, natural performance in greatly degraded for both adv and dec+adv in the CPSC data in Table \ref{table:results_china}. This degradation has been observed for adversarial training, and is likely due to the network's limited capacity, which forces a tradeoff between natural discriminative features and adversarially robust features \cite{yin_fourier_2019, ilyas_adversarial_2019}. Indeed, as the dimension of the input space increases, as is the case of the higher dimensional CPSC data, much larger networks and more training (both adversarial and natural) are needed to fit the robust but complex decision boundaries \cite{madry_towards_2019}, rendering adversarial training less feasible for high dimensional problems. Decorrelation alone also provides less benefit in this instance; we suspect that this is due to the increase in classes. As the number of classes increases, the final feature layer, where decorrelation takes place, must increase in dimension such that feature vectors extracted from different networks can be uncorrelated while still correlating with the correct class. As such, architectural changes may be needed to optimize the decorrelation mechanism in this instance. Despite this, unlike adversarial training, the combination of dec+part still provides increased accuracy against all PGD and SAP attacks without any degradation in natural accuracy, even outperforming dverge in most cases. Additionally, the $UA$, $R_{iu}$, and $R_{cc}$ for dec+part is superior to the baseline in all adversarial instances, and even outperforms adv in some instances. This suggests that dec+part can add adversarial robustness without sacrificing natural accuracy, even in higher dimensional problems.

While the metrics in Tables \ref{table:results} and \ref{table:results_china} overall indicate robustness benefits for linear feature decorrelation and Fourier partitioning, these results are admittedly heterogeneous. It is clear that especially in higher dimensional data, there is a strong tradeoff between natural performance and adversarial robustness. Furthermore, an ensemble's inference accuracy is not always correlated with the utility of its uncertainty estimation: some ensembles may pose better overall accuracy but worse $UA$, $R_{iu}$, and/or $R_{cc}$. It is important to evaluate BNNs in a method similar to how they could be deployed to observe potential tradeoffs between these qualities. This is the motivation behind the mixed dataset experiments (Figures \ref{fig:plot_curve_natural} and \ref{fig:plot_curve}), as it initially explores how well the uncertainty measures of different ensembles can prioritize clinician attention. Figure \ref{fig:plot_curve_natural} shows the worst performance on natural data with adversarially trained networks, reflecting the loss of natural performance with adversarial training. On the other hand, dec, part, and dec+part all perform comparably or better than the baseline and all other methods in the PhysioNet data; dec and dec+part also perform only marginally worse than the baseline in the natural CPSC 2018 data while showing superior performance on both perturbed datasets. Indeed, Figure \ref{fig:plot_curve} shows dec, part, and dec+part outperforming both dverge and the baseline of the PhysioNet data in this scenario, and dec+part achieves the best overall performance (smallest AUC) on the perturbed CPSC data.

Additionally, Figure \ref{fig:uncertainty_plots} illustrates that part and dec+part, and dec+adv are the methods that best maintain a higher uncertainty difference $\Delta I_{norm}$ across varying attacks magnitudes for both datasets. Interestingly, while adv and dec individually fail to maintain this uncertainty difference against stronger attacks on the CPSC data, the combination dec+adv performs better in this metric. Once again, we suspect that feature decorrelation may benefit from expansion of the feature space as the number of classes increases.

Our results generally show that the proposed modified linear feature decorrelation and Fourier partitioning methods show promise for diversifying extracted features in deep ensembles, and can be used in other high-dimensional classification problems, such as medical image analysis. Using the fast Fourier transform, Fourier partitioning is an efficient way to force ensemble models to extract different features. Previous work  mentioned several challenges with scaling decorrelated ensembles to larger problems, such as the need to train models in parallel and the large batch size needed to overdetermine the feature space with each training step\cite{wiedeman_disrupting_2022}. We find that our modifications, such as selecting the final hidden layer for decorrelation and compressing the feature space with random projections, has allowed us to scale this mechanism to higher dimensions (see details in Methods section).

Furthermore, we found that both methods added little to no extra training time, and penalized natural accuracy less than adversarial training. It should also be noted that both methods are orthogonal to gradient-based methods such as adversarial training and DVERGE, and thus can be combined with these methods.

A number of limitations exist for this study, which can be explored in future work:
\begin{itemize}
    \item \textbf{Optimization of the design space}: The introduction of Fourier partitioning and decorrelation introduce new hyperparameters for tuning. For decorrelation, the compression ratio for the features, dimension of the feature space, and batch size are all critical considerations. As previously discussed, we believe that expanding the feature space may be necessary as the number of discriminative classes increases. Fourier partitioning was inspired by the discovery that neural networks can often solve computer vision tasks with only partial frequency information \cite{yin_fourier_2019}. Thus, each ensemble filter should be designed to preserve sufficient information for the task but have non-overlapping vulnerabilities. Our experiments simply used two ‘ring filters’ which summed to an impulse response, but many other schemes could be explored in the same spirit. Additionally, these methods can be used in combination with DVERGE or adversarial training. Future experiments should extensively explore all these considerations, as well introduce new applications.
    \item \textbf{Investigation of robustness against various attacks, corruptions, and shifts}: This work focuses on adversarial attacks in ECG of varying magnitude, using both PGD and the more domain-specific SAP algorithms. However, robust uncertainty quantification is desirable in many other contexts, such as domain-specific noise corruptions, data domain shifts, and out-of-sample detection. Since neither decorrelation nor Fourier partitioning explicitly optimize against adversarial attacks, we hypothesize that their diversification benefits may extend to these other contexts. 
    \item \textbf{Experiments mimicking clinical deployment}: The tradeoffs between a model's inference accuracy and uncertainty accuracy make it necessary to test how deep models can work in synergy with clinicians. Our experiments compare the number of misclassified samples when uncertainty is used to theoretically prioritize clinician attention. In reality, however, clinical workflows are more complicated, and desirable outcomes will depend on the cost of false positives/negatives for different diagnoses, clinical resources, and the ability for human clinicians to more accurately diagnose certain diseases relative to a model. Thus, all these aspects should be considered in future translative work. Furthermore, we only tested a natural sample and an attacked sample where the distribution of attack magnitudes was roughly based on the assumption that larger perturbation attacks are be less common. Models should be rigorously tested specifically with the kinds and magnitudes of perturbations that one might expect in deployment. 
\end{itemize}

\section*{Conclusion}
Efficient and accurate confidence measurement is necessary for trust in AI systems. We have presented a novel approach for diverse network ensembles using two unique training methods which add little to no training time: a streamlined and accelerated decorrelation training strategy and a Fourier partitioning scheme. These ensembles achieve robustness by focusing on feature diversity between models. Additionally, we adapt adversarial training to ensembles, and test all methods along with DVERGE in the Bayesian ensemble framework. All approaches are applied to ECG classification with uncertainty estimation, and tested for stability against state-of-the-art adversarial ECG attacks, demonstrating their merits and potential in solving large problems. Incorrect diagnoses can cause major harm in many healthcare tasks where AI can work alongside clinicians; predicting model confidence in these contexts is crucial. Thus, we speculate that diverse ensembles will play a key role in elevating trustworthiness and confidence in AI for applications such as tomographic image processing, radiomics, and multimodal diagnosis. We see applications of this approach for robust uncertainty estimation with a diversified ensemble, which discourages different models from extracting redundant features. 

\section*{Methods}\label{sec:methods}
\subsection*{Ensemble Training and Inference}
Basic ensembles consist of multiple deep neural networks trained for the same task. An ensemble inference on sample $x$ is simply the average output of each model in the ensemble. Thus, for an ensemble with  models $f_1, f_2, ... f_K$:
\begin{align*}
\hat{y} = f_{ens}(x) &= \frac{1}{K} \sum_{k=1}^K f_k(x)
\end{align*}

Note that for a classification task, this output is a discrete probability distribution.

For estimating epistemic uncertainty, we adopt the approach from \cite{mobiny_2021}, which defines the uncertainty of sample $x$ as the mutual information between the inferred label $\hat{y}$ and the underlying parameter distribution. In other words, how much additional information sample $x$ tells us about the true parameters:

\begin{align*}
I(y,\theta|x,\mathcal{D}) = H(y|x, \mathcal{D}) - H(y|x,\theta, \mathcal{D}) = H(y|x, \mathcal{D}) - \mathbb{E}_{\theta|\mathcal{D
}}[H(y|x,\theta)]
\end{align*}

$\mathcal{D}$ is the training data. The first term is intractable, but can be estimated using the network ensemble as the entropy of the expected inference \cite{mobiny_2021}. Thus, for ensemble with $f_1, f_2, ... f_K$ models, each of which output a discrete probability distribution over $C$ classes:

\begin{align*}
I(y,\theta|x,\mathcal{D}) &= - \sum_{c=1}^C f_{ens}(x)[c] \text{log} f_{ens}(x)[c]  + \frac{1}{K} \sum_{k=1}^K \sum_{c=1}^C f_k(x)[c] \text{log} f_k(x)[c] \\
\end{align*}

The scale of $I$ is relative and can vary between models and ensembles. Thus, we normalize the uncertainty with the minimum and maximum uncertainty values found during training.

\begin{align*}
I_{norm} = \frac{I-I_{min}}{I_{max} - I_{min}}
\end{align*}

Note that test samples can have greater or less uncertainty than any sample encountered in the training set. Thus, values for $I_{norm}$ are not necessarily limited to $[0,1]$. For a threshold $I_T$ which classifies samples as either 'certain' or 'uncertain', metrics $R_{cc}$, $R_{iu}$, and $UA$ were calculated empirically over an adversarial dataset based on the number of correct \& certain, correct \& uncertain, incorrect \& certain, and incorrect \& uncertain samples.

For the baseline ensemble, each model was trained independently in sequence. Other ensembles were trained using the methods described below.

\subsection*{Decorrelation training}
\subsubsection*{Decorelation of Two Networks}
The intent of decorrelation training between two networks is to minimize the Pearson correlation coefficient between the latent features extracted by the two networks, as explained in \cite{wiedeman_disrupting_2022}. Minimizing this value incentivises the networks to extract different features for a task, reducing the transferability of network weaknesses \cite{wiedeman_disrupting_2022, ilyas_adversarial_2019}. Assume two classification models $f_1$ and $f_2$, and sample batch $X$ of inputs with accompanying labels $Y$. Next, $Z_i = f_i^l(X)$ for $i=1,2$ are the latent feature extracted by model $i$ at layer $l$ from batch $X$. Note that $Z_i \in \mathbb{R}^{N \times D}$ where $N$ is the batch size and $D$ is the dimension of the latent space. A linear relationship estimating $Z_2$ from $Z_1$ with weights $W$ can be found using ordinary least squares regression:

\begin{align*}
\underset{W}{\operatorname{minimize}} ||&Z_2 - \mathbf{Z_1}W||^2_2 \\
\text{where} \;\; \mathbf{Z_1} &= [Z_1, \mathbf{1}] \\
\underset{W}{\operatorname{min}} ||Z_2 - \mathbf{Z_1}W||^2_2 &= 
||Z_2 - \mathbf{Z_1}(\mathbf{Z_1}^\top \mathbf{Z_1})^{-1}\mathbf{Z_1}^\top) Z_2||^2_2 = SS_{res} \\ \nonumber
\end{align*}

The Pearson correlation coefficient is then the ratio between the residual and total sum of squares: 

\begin{align*}
R^2 = 1 - \frac{SS_{res}}{SS_{total}} &= 1 - \frac{||(I - \mathbf{Z_1}
(\mathbf{Z_1}^\top \mathbf{Z_1})^{-1}\mathbf{Z_1}^\top) Z_2||^2_2}
{||Z_2 - \bar{Z_2}||^2_2} \nonumber
\end{align*}

To reduce this term during training, the decorrelation loss is defined as:

\begin{align}
\mathcal{L}_R &= \text{log}(SS_{total} + \epsilon) - \text{log}(SS_{res} + \epsilon) \\
\mathcal{L}_R (Z_1, Z_2) &= \text{log}(||Z_2 - \bar{Z_2}||^2_2 + \epsilon) - \text{log}||(I - \mathbf{Z_1}
(\mathbf{Z_1}^\top \mathbf{Z_1})^{-1}\mathbf{Z_1}^\top) Z_2||^2_2 + \epsilon) \label{eqn:LR}
\end{align}

where $\epsilon$ is some small constant for stability (set to $10^{-5}$ in our experiments) and $\bar{Z_2}$ is an $N \times D$ matrix where each row is the sample mean of $Z_2$. This decorrelation can be applied to model training simply by weighting and adding this loss to a conventional training objective (e.g., cross-entropy loss) for both networks, balancing feature decorrelation and individual network performance.

\subsubsection*{Scaling Decorrelation to Ensembles}
While decorrelation of two networks was shown to reduce transferability of adversarial attacks in \cite{wiedeman_disrupting_2022}, several issues impede its use in larger ensembles and higher-dimensional problems. First, this method requires training multiple networks in parallel to obtain $Z_1$ and $Z_2$, which linearly scales the memory needed for training, and quickly becomes infeasible for more networks, larger networks, and larger data sizes. Secondly, calculating the loss in Equation \ref{eqn:LR} requires taking the pseudo-inverse of $\mathbf{Z_1} \in \mathbb{R}^{N \times D+1}$, requiring batch size $N>D+1$ for a sufficiently overdetermined system. Higher dimensional problems often necessitate higher dimensional latent spaces and smaller batch sizes, rendering this condition impractical.

To solve the above problems and extend the method to ensembles consisting of more than two networks, we employ the following modifications:

\begin{itemize}
\item Selecting the regression layer: Most deep classification networks can be divided into a multi-layer feature extractor and a linear layer that followed by a softmax function (i.e., a logistic regression). We select the network layer just before this final linear layer, as it represents the highest-level features, and is typically lower dimension than previous layers.
\item Dimensionality reduction via random projections: Prior to computing the pseudo-inverse of the regressor, we compress its $D+1$ dimensionality to $r<D+1$ by applying a random projection $R \in \mathbb{R}^{D+1 \times r}$. To balance the asymmetry in this relationship, we randomly select which network's extracted features act as the regressor and which as the regressand with each training batch. Our new loss is expressed as follows:

\begin{align*}
\mathcal{L}^*_R(Z_1, Z_2) &=
        \left\{ \begin{array}{ll}
            \mathcal{L}_R (Z_1, Z_2R) & \text{with prob. 0.5} \\
            \mathcal{L}_R (Z_2, Z_1R) & \text{with prob. 0.5}
        \end{array} \right. \\
        R &\in \mathbb{R}^{D+1 \times r} \sim N(0,1/\sqrt{D})
\end{align*}

Although these projections individually do not capture all information in the regressor, they are drawn randomly with each training batch, preventing the networks from only decorrelating a subspace of the original feature space.

\item Models are trained in sequence instead of parallel: The first network is trained without any decorrelation loss. After training a model, its extracted features on all training samples are saved. While training the next model, these features are loaded with the corresponding batch samples, and then used for decorrelation. As such rather than dynamically decorrelating multiple networks at once, which requires simultaneous training of all networks, we simply use the features extracted by the previously trained networks as constant values to decorrelate against. For decorrelating against multiple models, we average the modified correlation loss against all the previously trained models. Thus, the entire decorrelation loss for model $k$ in an ensemble:

\begin{align}
\mathcal{L}_{cor}(Z_k, Z_{k-1} \cdots Z_0) &= 
\frac{1}{k}\sum_{i=0}^{k-1} \mathcal{L}^*_R(Z_k,Z_i) \label{eqn:Lcor}
\end{align}
\end{itemize}

\subsubsection*{Final Decorrelation Scheme}
\begin{figure*}
\centering
\includegraphics[width=16.0cm]{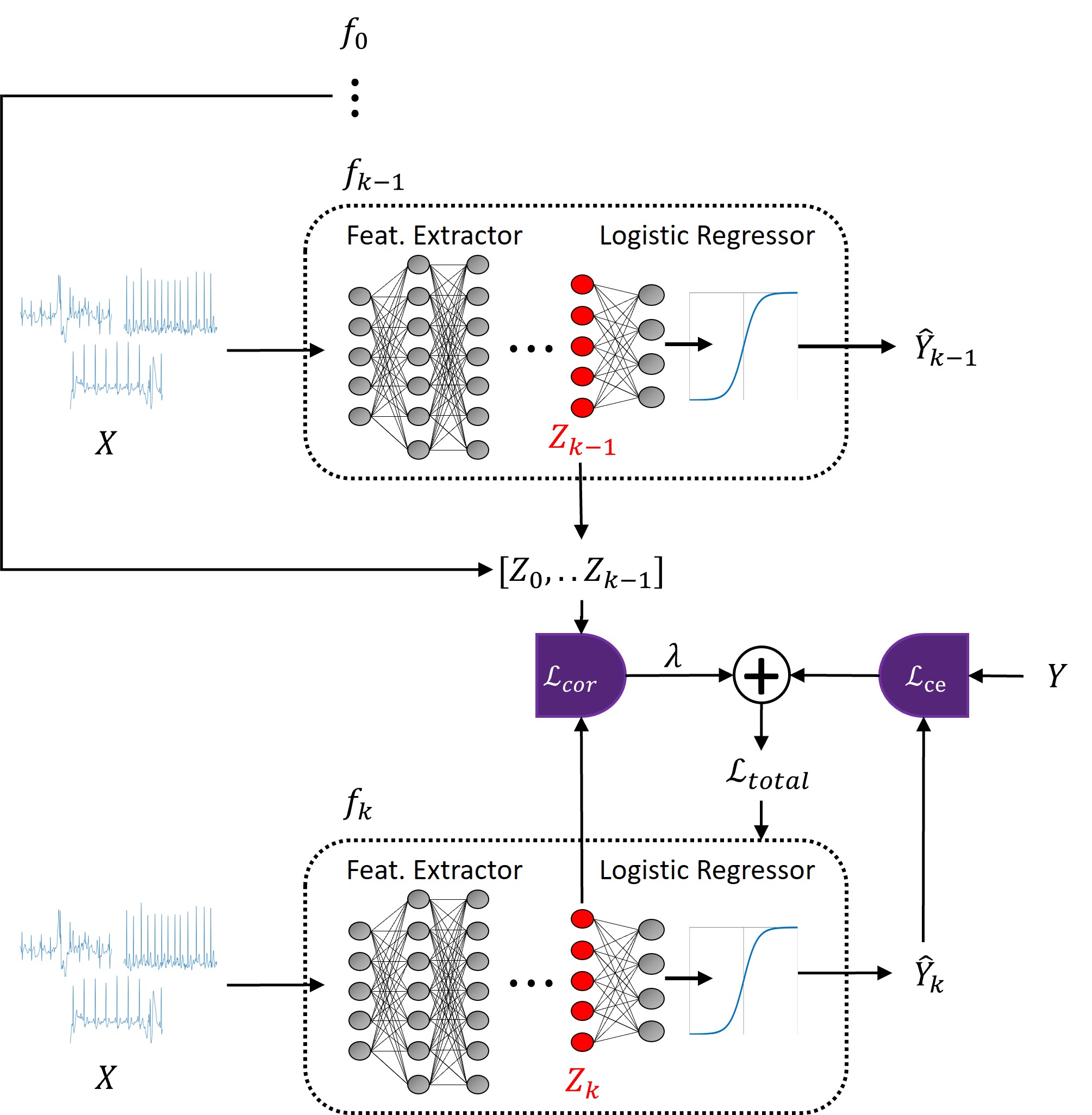}
\caption{Illustration of the decorrelation training process. The current model $f_k$ is trained using both cross entropy and a correlation loss. The correlation loss references previous models’ extracted sample features as opposed to training multiple models in parallel.}\label{fig:dec_diagram}
\end{figure*}

Figure \ref{fig:dec_diagram} illustrates the sequential training of the decorrelated ensemble. The total loss for model $k$ on batch $(X_b, Y_b)$ with extracted features from $X_b$ using $k-1$ previous models as $(Z_{k-1,b}, ... Z_{0,b})$ is:

\begin{align}
\mathcal{L}_{total} &= \mathcal{L}_{ce}(f_k(X_b),Y_b) + \lambda \mathcal{L}_{cor}(Z_{k,b} Z_{k-1,b} \cdots Z_{0,b}) 
\label{eqn:Ltotal}
\end{align}

where $\mathcal{L}_{ce}$ is the cross-entropy loss and $\lambda$, a weighting hyperparameter. The implementation is summarized in the Algorithm \ref{alg:dec_alg}. While stochastic gradient descent is shown here, any optimizer can be used for the gradient step.

\begin{algorithm}
\caption{Training Step for Model $f_k$ (with parameters $\theta_k$) using decorrelation. $\lambda$ and $r$ are hyperparameters.}\label{alg:dec_alg}
\begin{algorithmic}
\State Draw $X_b, Y_b, [Z_{k-1,b} \cdots Z_{0,b}]$ \Comment{Draw training batch and corresponding features from prior models}
\State $Z_{k,b} \gets f^l_k(X_b)$
\State $N, D \gets shape(Z_{k,b})$
\State $\hat{Y_b} \gets f_k(X_b)$
\State $\mathcal{L} \gets \mathcal{L}_{ce}(\hat{Y_b}, Y_b)$
\State $i \gets 0$
\While{$i \leq k-1$}
   \State $Z_1, Z_2 \gets Z_{k,b}, Z_{i,b}$
    \If {$t \sim Uniform[0,1] < 0.5$}
        $Z_1, Z_2 \gets Z_2, Z_1$
 \EndIf
\State $R \sim N(0,1/\sqrt{D}) \in \mathbb{R}^{D+1 \times r}$
\State $Z_1 \gets [Z_1, \mathbf{1}]R$
\State $\mathcal{L} \gets \mathcal{L} + \frac{\lambda}{k} \mathcal{L}_R(Z_1, Z_2)$ \Comment{Apply decorrelation loss from Equation \ref{eqn:LR}}
\State $i \gets i+1$
\EndWhile

\State $\theta_k \gets \theta_k - \eta \nabla_{\mathcal{L}} \theta_k$
\end{algorithmic}
\end{algorithm}

\subsection*{Fourier Partitioning Scheme}
\begin{figure*}
\centering
\includegraphics[width=14.0cm]{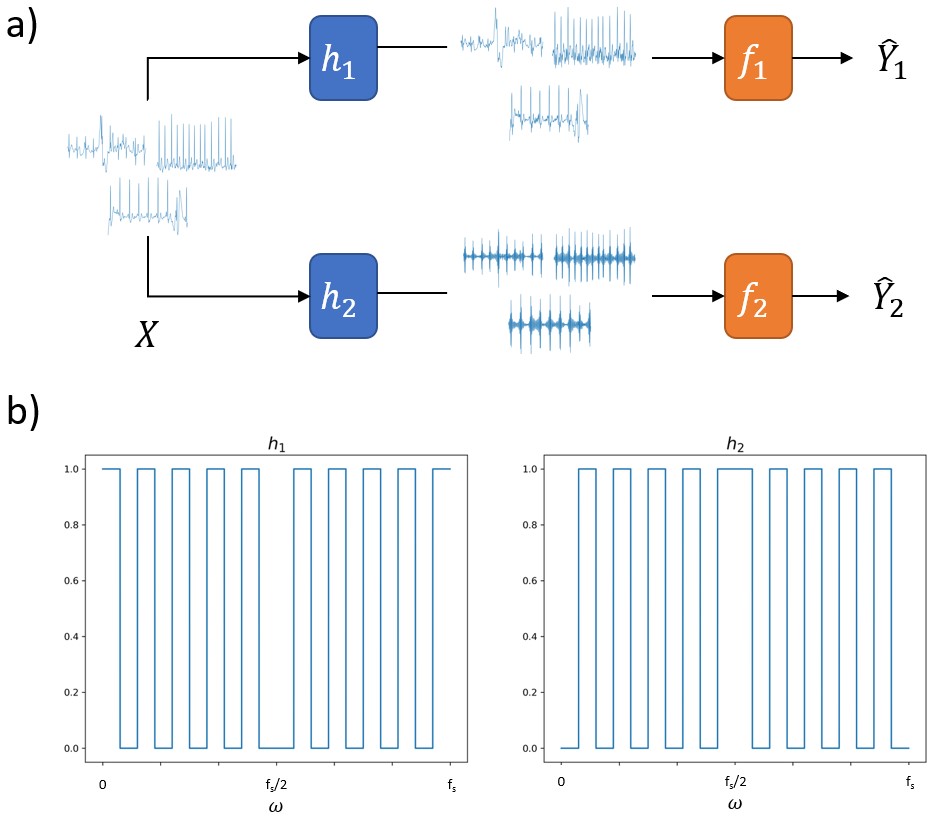}
\caption{Illustration of Fourier transform-based input data decomposition. a) Diagram showing frequency partitioning each sample into two inputs, which are fed into different models, where h is a partitioning filter, and $f$ is a classification model; and b) the frequency responses of $h_1$ and $h_2$ (real-valued only). $f_s$ is the data sampling frequency (300Hz for PhysioNet 2017, 500Hz for CPSC 2018).}\label{fig:filter_diagram}
\end{figure*}

With the Fourier partitioning scheme, the first ensemble model has no modification. The other models were trained normally but inputs were filtered during both training and inference (Figure \ref{fig:filter_diagram}). This approach is inspired by \cite{yin_fourier_2019}, which showed that neural networks can achieve high classification accuracy in many computer vision tasks with only a portion of an input's frequency data, and that models often overfit to discriminative features in specific frequency bands.

\begin{align*}
\hat{y}_{i,k} &= f_k(h_k \circledast x_i)
\end{align*}

In practice, filter convolution was done by pointwise multiplication in the Fourier domain, computed using the fast Fourier transform.

\subsection*{Adversarial Attacks}
Adversarial attacks are formulated by maximizing the loss objective $L$ of the model f by modifying $x$ (with the paired label $y$) within a set of valid perturbations $\Delta$:

\begin{align}
& \underset{\delta}{\text{maximize}} \;\; J(x+\delta, y) \label{eqn:objective} \\
& \text{subject to} \;\;\; \delta \in \Delta(x) \nonumber
\end{align}

Two algorithms were used to craft adversarial attacks: projected gradient descent (PGD) and smoothed adversarial perturbations (SAP). PGD is widely used as a strong attack with an $\ell_\infty$ bound through iterative optimization \cite{madry_towards_2019}:

\begin{align}
x'_i = \text{Clip}_\varepsilon(x'_{i-1} + \alpha \text{sgn}(\nabla_x L(f(x'_{i-1}),y))) \label{eqn:pgd}
\end{align}

where $x'_i$ is the sample at the $i^{th}$ iteration, $y$ is the corresponding label, $\alpha$ is a step size, $L$ is the loss function, and the clipping operation clips all values to be within the $\ell_\infty$ ball of radius $\varepsilon$ around $x$, as well as any implicit bounds on the domain of $X$.

SAP is a variation of PGD designed to craft smooth attacks for ECG signals. The details of SAP can be found in \cite{han_deep_2020}. To summarize, perturbation $\theta$ is iteratively optimized in a fashion similar to PGD, but is also convolved with a sequence of $M$ Gaussian kernels at every step, each of which are parameterized by their width s and standard deviation $\sigma$:
\begin{align}
\theta_i &= \text{Clip}_\varepsilon(\theta'_{i-1} + \alpha \text{sgn}(\nabla_\theta L(f(x'(\theta_{i-1})),y))) \nonumber \\
x'(\theta) &= x + \frac{1}{M} \sum_{m=1}^{M} \theta \circledast K(s_m, \sigma_m) \label{eqn:sap}
\end{align}
The convolution with Gaussian kernels smooths high frequency perturbations, removing unrealistic square wave artifacts. PGD and SAP attacks were optimized over 20 steps in total. For both attacks, $\alpha$ was scaled as $\varepsilon / 10$. All adversarial attacks were crafted from the validation to target one of the models in an ensemble.
For experiments with PhysioNet 2017 data, the convolution kernels are identical to those used in \cite{han_deep_2020}: $s = (5, 7, 11, 15, 19), \sigma=(1, 3, 5, 7, 10)$. For the CPSC 2018 experiments, we used  $s = (9, 11, 15, 19, 21), \sigma=(5, 7, 10, 13, 17)$.

\subsection*{Experimental Details}
The two ECG datasets used in this work are from the PhysioNet 2017 and CPSC 2018 challenges. The PhysioNet dataset contains 30-60 second duration single channel ECG recordings sampled at 300Hz. All samples were were zero-padded to be 60 seconds. Class labels are {Normal, A. Fib., Other Rhythm, Noise}. Scaling of the signal magnitudes were identical to those in \cite{han_deep_2020}. The CPSC 2018 data contains 6-60 second ECG recordings at 500Hz with nine class labels: {Normal, A. Fib, 1\textsuperscript{st}-Degree Atrioventricular Block, Left Bundle Branch Block, Right Bundle Branch Block, Premature Atrial Contractions, Premature Ventricular Contraction, ST-segment depression, ST-segment elevated}. All samples were truncated or zero-padded to be 48 seconds. For simplicity, the minority of samples labelled with multiple diagnoses were not used. Additionally, all channels were normalized from -1 to 1.

For all experiments, each ensemble consisted of $K=3$ classification networks, each with the architecture in used \cite{goodfellow_towards_2018} for experiments using the PhysioNet 2017 data. For experiments using the CPSC 2018 data, this architecture was simply modified to have 12 input channels and 9 output classes. Each network was trained for 80 epochs (batch size of 64) using the Adam optimizer with a learning rate of $10^{-3}$. Pytorch 1.8.1 was used with two NVIDIA Titan RTX GPUs, and a 90/10 training/validation split. For all ensembles using decorrelation training, hyperparameters $r=32$ and $\lambda=0.2$ were used (the uncompressed latent dimension $D$ was 64).

For ensembles that underwent additional ensemble adversarial training, each model in the ensemble was sequentially trained using adversarial samples. In practice, this is identical to regular training, except each sample batch is perturbed using PGD (Equation \ref{eqn:pgd}) prior to the forward training step. The perturbations target the model being trained, and use $\varepsilon = 10$  and $0.025$ for the PhysioNet 2017 and CPSC 2018, respectively. Each model is trained for an additional two hours, translating to six extra training hours total for each adversarially trained ensemble.

DVERGE training is similar to adversarial ensemble training, in that both use PGD to perturb a sample $x_s$ within a small $\ell_\infty$ bound. Differences are that in DVERGE 1) this optimization is used to maxmize similarity between the distilled features of $x_s$ and some other randomly drawn sample $x$ at some randomly selected feature layer $l$  of the network, and 2) the samples perturbed using network $i$ are used to train \emph{other} networks $i \neq j$. Thus, we use the feature distillation objective and training procedure from \cite{yang_dverge_2020}:
\begin{align}
x'_{f^l_i}(x,x_s) &= \underset{z}{\text{argmin}} ||f^l_i(z) - f^l_i(x)||^2_2, \\
\text{subject to} \;\; \varepsilon &\geq ||z-x_s||_\infty
\nonumber
\label{eqn:dverge}
\end{align}
Values for $\varepsilon$ were identical to those used in adversarial ensemble training. The DVERGE training also ran for the same total training time (6 additional hours) as adversarial ensemble training. For each batch, feature layer $l$ was randomly, uniformly selected from the post batch-norm layers of the networks.  

\section*{Data Availability}
Data used in this paper is from the 2017 PhysioNet Cardiology Challenge \cite{clifford_af_2017} and the 2018 China Physiological Signal Challenge  \cite{china_2018_data}. Code for implementing our methods and replicating experiments can be found at: https://github.com/WANG-AXIS/DNA\_ECG.

\bibliographystyle{unsrt}
\bibliography{main} 

\section*{Acknowledgments}
This work was partially supported by U.S. National Institute of Health (NIH) grants R01EB026646, R01CA233888, R01CA237267, R01HL151561, R21CA264772, R01EB031102, and National Science Foundation Graduate Research Fellowship supporting C.W.

\section*{Author Contributions}
C.W. and G.W. jointly conceived the idea for this study. C.W. designed code for executing all experiments and drafted the paper. G.W. was heavily involved in supervising the project, interpreting results, and editing the paper.

\section*{Competing Interests}
The authors declare no competing interests.

\end{document}